\documentclass[final,5p,times,twocolumn]{elsarticle}

\usepackage{lineno}
\usepackage{graphicx}
\usepackage{amssymb}
\usepackage{pdflscape}
\usepackage[english]{babel}
\usepackage[dvipsnames,svgnames,x11names]{xcolor}
\usepackage{booktabs,tabularx}
\usepackage{multirow}
\usepackage{subfig} 
\usepackage{hyperref}
\usepackage{amsmath}
\usepackage{commath}
\usepackage[english]{babel}
\usepackage[ruled]{algorithm2e}
\SetEndCharOfAlgoLine{}
\usepackage[leftmargin=6em,rightmargin=12em,indentfirst=false]{quoting}
\usepackage{siunitx}

\usepackage[final]{changes}
\usepackage{float}

\usepackage{lineno}
\usepackage{tikz}
\usepackage[export]{adjustbox}

\usepackage{graphicx}
\usepackage[utf8]{inputenc}
\usepackage[export]{adjustbox}
\usepackage{wrapfig}
\usepackage{dcolumn}

\makeatletter
\newcolumntype{T}[3]{>{\textfont0=\the@{#1}{#2}{#3}}c<{\DC@end}}
\makeatother

\usepackage{pgfplots}
\pgfplotsset{width=10cm,compat=1.9}

\usepackage{array}

\newcolumntype{L}[1]{>{\raggedright\let\newline\\\arraybackslash\hspace{0pt}}m{#1}}
\newcolumntype{C}[1]{>{\centering\let\newline\\\arraybackslash\hspace{0pt}}m{#1}}
\newcolumntype{R}[1]{>{\raggedleft\let\newline\\\arraybackslash\hspace{0pt}}m{#1}}

\usepackage{subfiles}

\usepackage[utf8]{inputenc}
\usepackage[T1]{fontenc}
\usepackage{lineno}
\usepackage{amsmath}
\usepackage{gensymb}
\usepackage{threeparttable}
\usepackage{graphicx}
\usepackage{multirow}
\usepackage{makecell}
\usepackage[nameinlink,capitalise]{cleveref}

\usepackage{todonotes}

\journal{Scientific Data}
\begin{document}
	
\begin{frontmatter}

\title{District-scale surface temperatures generated from high-resolution longitudinal thermal infrared images}

\author{Subin Lin$^{1}$, Vasantha Ramani$^{1}$, Miguel Martin$^{1}$, Pandarasamy Arjunan$^{1,2}$, Adrian Chong$^{2}$, Filip Biljecki$^{4,5}$, Marcel Ignatius$^{4}$, Kameshwar Poolla$^{3}$, Clayton Miller$^{2}{*}$}

\address{$^{1}$Berkeley Education Alliance for Research in Singapore, CREATE Tower 1 Create 6 Way, 138602, Singapore}
\address{$^{2}$Robert Bosch Centre for Cyber-physical Systems, Indian Institute of Science, Bengaluru, Karnataka, 560012, India}
\address{$^{3}$Department of the Built Environment, College of Design and Engineering, National University of Singapore, 4 Architecture Drive, 117566, Singapore}
\address{$^{4}$Department of Electrical Engineering and Computer Sciences, University of California, Berkeley, CA, USA}
\address{$^{5}$Department of Architecture, College of Design and Engineering, National University of Singapore, 4 Architecture Drive, 117566, Singapore}
\address{$^{6}$Department of Real Estate, Business School, National University of Singapore, 15 Kent Ridge Drive, 119245, Singapore}
\address{$^*$Corresponding Author: clayton@nus.edus.sg, +65 81602452}

\begin{abstract}
This paper describes a dataset collected by infrared thermography, a non-contact, non-intrusive technique to acquire data and analyze the built environment in various aspects. While most studies focus on the city and building scales, an observatory installed on a rooftop provides high temporal and spatial resolution observations with dynamic interactions on the district scale. The rooftop infrared thermography observatory with a multi-modal platform capable of assessing a wide range of dynamic processes in urban systems was deployed in Singapore. It was placed on the top of two buildings that overlook the outdoor context of the National University of Singapore campus. The platform collects remote sensing data from tropical areas on a temporal scale, allowing users to determine the temperature trend of individual features such as buildings, roads, and vegetation. The dataset includes 1,365,921 thermal images collected on average at approximately 10-second intervals from two locations during ten months. 
\end{abstract}


\begin{keyword}

Infrared thermography observatory\sep Built environment \sep IR dataset

\end{keyword}
\end{frontmatter}


\section{Background \& Summary}
\label{sec:background_summary}
Urban ecosystems are the biggest, most dynamic, and most complicated man-made systems, with millions of people interacting and hundreds of governing agencies \cite{dobler2021urban,2022_jag_geoai}. Urban modernization has encouraged a significant portion of the global population to move to urban regions.
Consequently, the built environment has grown rapidly, and building energy consumption has increased, which augments the release of CO$_2$ \cite{MSS:ieabuilding:2022,miguel2021physically}. Improving energy efficiency has been urged by various sectors \cite{lucchi2018applications}. The scientific community has exerted significant efforts to enhance its understanding of the built environment by leveraging a wide range of sensing technologies \cite{miguel2021physically}. Furthermore, the advent of computational techniques has revolutionized data collection and analysis, enabling us to process vast quantities of information. \cite{dobler2021urban}. 

Infrared thermography has been widely used in the built environment for many research purposes, such as urban heat island \cite{ngie2014assessment, almeida2021study}, building diagnostics \cite{balaras2002infrared,lucchi2018applications,2021_buildsys_acir}, and urban heat fluxes \cite{sham2012verification}. Infrared thermography can provide images showing the surface temperature of different elements in the built environment with a lower cost and effort compared to similar kinds of sensor networks \cite{martin2022infrared}. It detects infrared electromagnetic radiation emitted by the inspected objects and is commonly used for non-destructive testing and non-contact diagnostic technology \cite{balaras2002infrared, lucchi2018applications}. It is based on an infrared imaging system calibrated to measure surfaces' emissive power distribution at various temperature ranges. The infrared camera generates a series of two-dimensional, readable thermal images, with different colors and tones representing different temperatures \cite{balaras2002infrared, lucchi2018applications}. Infrared systems have been leveraged to analyze the built environment across various scales: Satellites, capable of wide-ranging data collection, are primarily used for city-scale analyses. This typically covers entire cities or large urban regions. Aerial vehicles, possessing a more focused reach, are frequently employed for investigations at both city and district scales. The term \emph{district scale} indicates a specific urban district or neighborhood. It refers to a geographical area larger than a single building or block but smaller than an entire city or region. This level of analysis allows for considering the collective impact of multiple buildings and infrastructure elements within a defined urban district. Due to their unique vantage point, rooftop observatories are particularly effective for district-scale studies. These offer a detailed view of the urban microclimate within a specific district or neighborhood. Lastly, drones and handheld devices are used for detailed investigations at smaller scales. Their portability and close-range capabilities make them particularly suited for analyses at the building scale (focusing on individual buildings) and human scale (focusing on the immediate surroundings of individuals or small groups)\cite{martin2022infrared}. Many previous and current studies focus on the city and building scales and could not show the dynamic interactions on the district scale with high-resolution \cite{miguel2021physically, dobler2021urban}. Furthermore, urban climate studies have been focusing on the temperate climate zones, especially in North America and Europe, and there is a lack of such studies focusing on the tropical climate zones \cite{chew2021interaction, roth2007review, 2022_3dgeoinfo_lcz_sg_svi}. 

This work introduces the first thermal observatory deployed in Singapore, a tropical city, with a multi-modal platform with flexibility and high resolution to assess a wide range of dynamic processes in urban systems. The thermal observatory was set up on the roof of buildings that overlooked several educational buildings on the National University of Singapore campus. The platform gathers remote sensing data from a tropical area at a high temporal resolution at approximately 10-second intervals, providing the temperature trends of specific urban elements such as buildings, roads, and vegetation. Demonstration codes were given with data preprocessing, such as segmentation, in order to handle and analyze the generated raw data and allow users to use the data with high flexibility, catering to their research needs.

\subsection*{Thermal images data collection}
\begin{figure*}[h!]
\centering
\includegraphics[width=\linewidth]{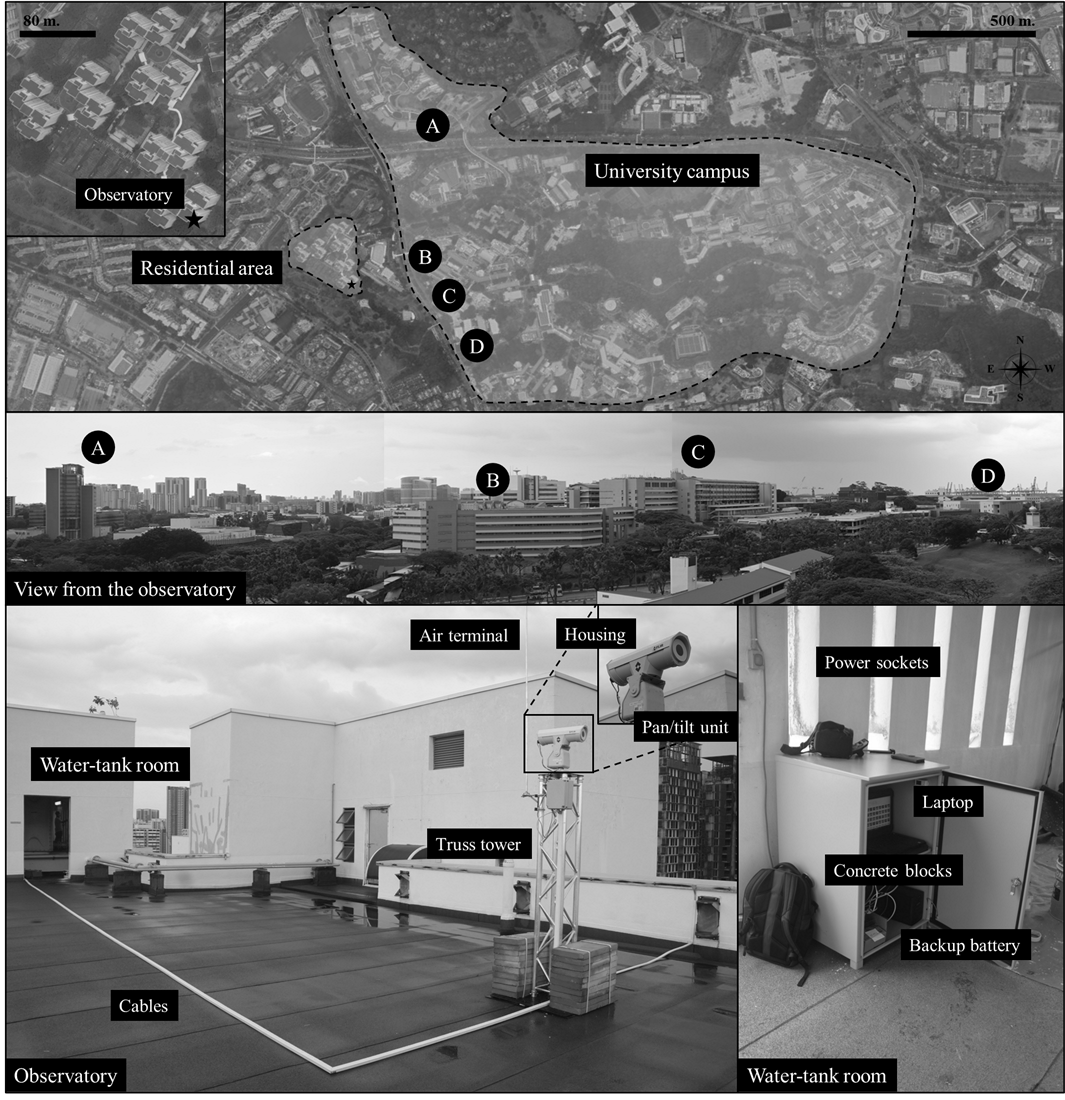}
\caption{Observatory installed on the rooftop of the residential area in Singapore and its captured information. Source of the imagery: Google Earth. \cite{https://doi.org/10.48550/arxiv.2210.11663}}
\label{fig:position_obser}
\end{figure*}

Thermal images are taken from a FLIR A300 (9 Hz) thermal camera atop an urban-scale IR observatory, as shown in Figure \ref{fig:position_obser}. The detailed specifications of the thermal camera are listed in Table \ref{tab:camera}. The thermal camera is protected by a housing against rains with IP67 protection. It was attached to a pan/tilt device that can spin 360 \degree along its horizontal axis to record thermal images in multiple locations. These structures (thermal camera, housing, pan/tilt) were placed on a 2-m-high truss tower to avoid impediments while taking thermal images. Concrete blocks serve as a stabilizing support, and an air terminal serves as lightning protection for the truss tower. Two plugs were added to the truss tower to allow a backup battery in the water tank room to power both the thermal camera and the motor of the pan/tilt device. The backup battery is continually charged from the building's electrical supply to enable the pan/tilt unit and thermal camera to be operational for up to two hours during a power outage. A laptop with a weatherproof casing in the water tank room was also linked to the pan/tilt device and thermal camera for customizing and reviewing the collection of thermal images.

\begin{figure*}[h!]
\centering
\includegraphics[width=\linewidth]{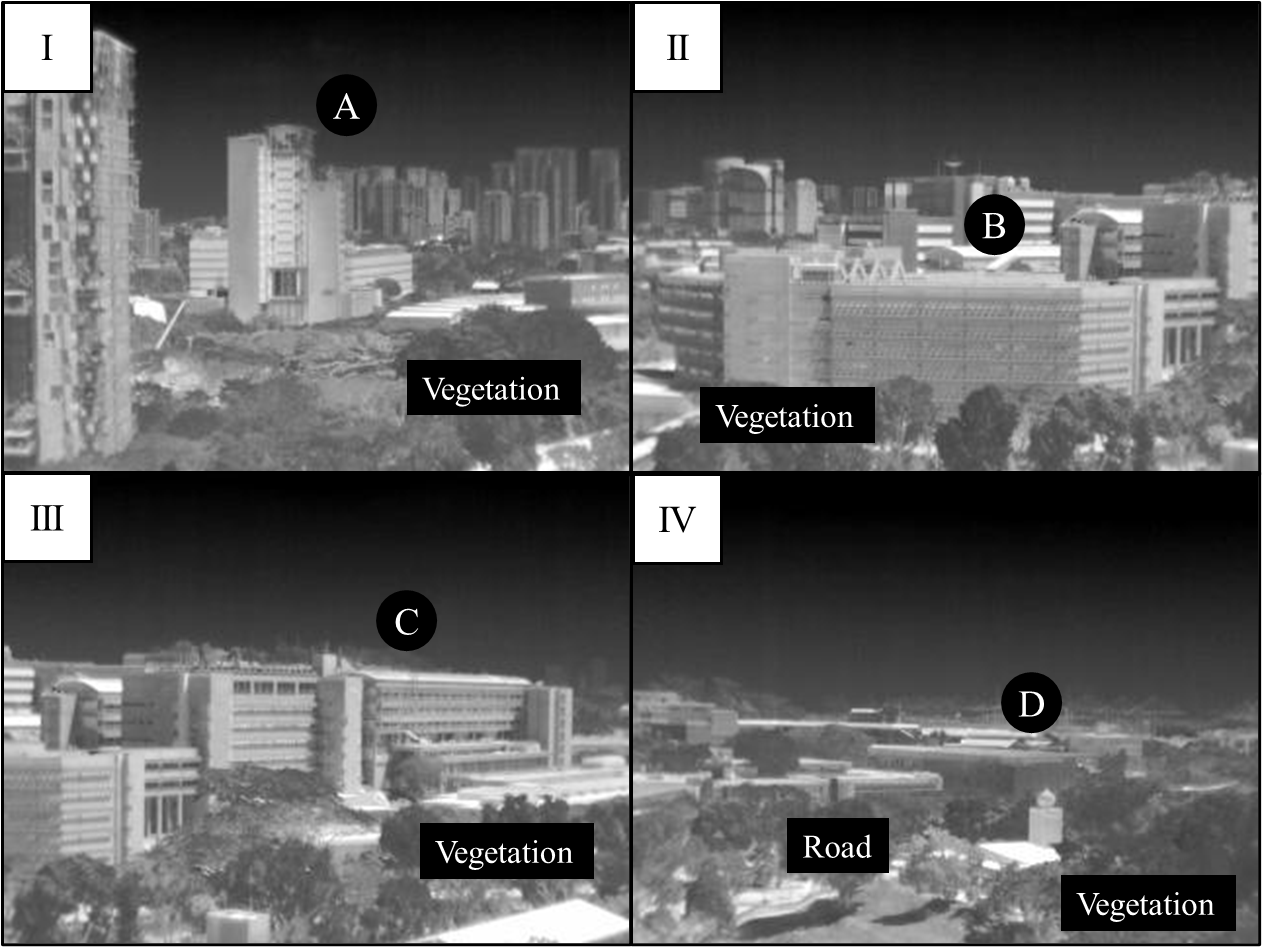}
\caption{Positions where the observatory captured thermal images on Kent Vale \cite{https://doi.org/10.48550/arxiv.2210.11663}}
\label{fig:position_abcd}
\end{figure*}

The thermal images are collected from two separate locations, namely Kent Vale and S16.
The Kent Vale observatory is located on the rooftop of a 42-m-tall building in a residential area, which is located in front of a university campus consisting of office and educational buildings, in order to access the district-scale analysis. The pan/tilt unit was configured from two separate software to allow the thermal camera to take images from four positions (I, II, III, and IV), as shown in Figure \ref{fig:position_abcd}. The first software was installed in a video encoder to control the positions of the pan/tilt unit to take a thermal image. The second software developed by NAX Instruments Pte Ltd was installed in the laptop to control the moment to take a thermal image, which could then be saved either in JPEG or FFF file format. The observatory can capture thermal images of four buildings, as shown in Figure \ref{fig:position_obser}. Building A, known as CREATE, is centrally located at Position I and at a height of 68 meters. Curtain walls cover a significant portion of its facade. Buildings B (E1A) and C (EA) are each centered at positions II and III, respectively. Both buildings are roughly 27 meters tall, characterized by their concrete walls and single-pane windows. Building D, or SDE4, can be observed at Position IV. It is structured with metal grids mounted on a concrete frame and is designed to be a net-zero building. It stands at a height of approximately 24 meters. Surface temperatures of trees are also observed from the above positions. A road with traffic exists in front of Buildings B, C, and D, and the road can be observed in thermal images taken from Position IV. The observatory moved from Position I to IV sequentially and recorded thermal images at various intervals. The detailed time for the image taken is shown in its file name. Then, using a 4G Internet connection fitted on the laptop, the captured thermal images are stored on the Google Drive repository.

\begin{table}[ht]
\centering
\begin{tabular}{|l|l|}
\hline
Image resolution & 16 bit, 320 $\times$ 240 pixels \\
\hline
Sensor & Uncooled Microbolometer FPA \\
\hline
Thermal sensitivity & 50 mK @ 30 \degree C \\
\hline
Spectral range & 7.5 - 13 $\mu$m \\
\hline
Field of view (FOV) & 25\degree (H) and 18.8\degree (V) \\
\hline
Accuracy & ±2\degree or 2\% of reading \\
\hline
Camera Weight & 0.7 kg \\
\hline
Power supply & 110/220 V AC \\
\hline
Size & 170mm $\times$ 70mm $\times$ 70mm \\
\hline
\end{tabular}
\caption{\label{tab:camera} Information about the FLIR A300 thermal camera}
\end{table}

\begin{figure*}[h!]
\centering
\includegraphics[width=\linewidth]{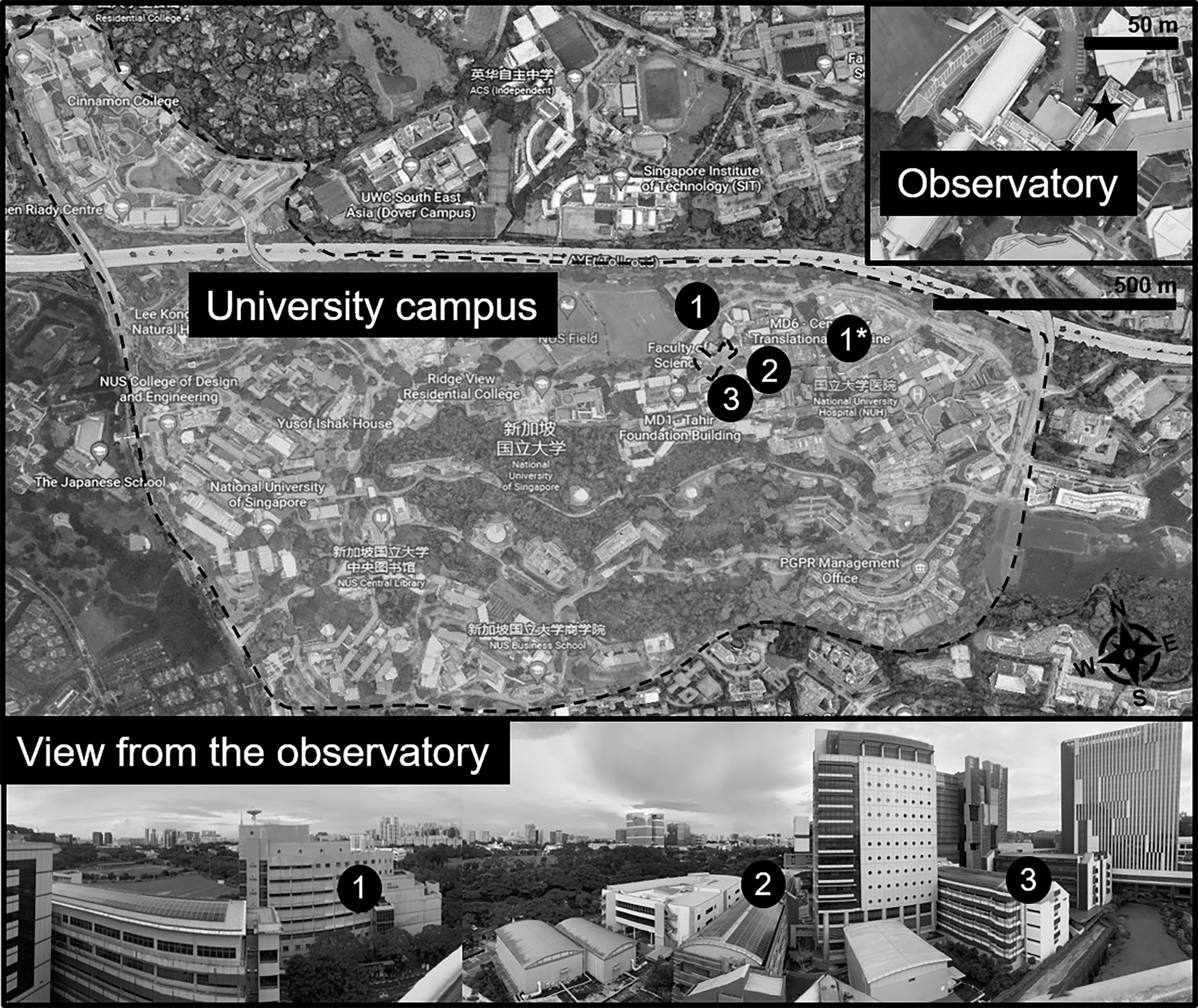}
\caption{Observatory installed on the rooftop of the university campus in Singapore and its captured information. Source of the imagery: Google Earth.
}
\label{fig:position_obser_s16}
\end{figure*}

\begin{figure*}[h!]
\centering
\includegraphics[width=\linewidth]{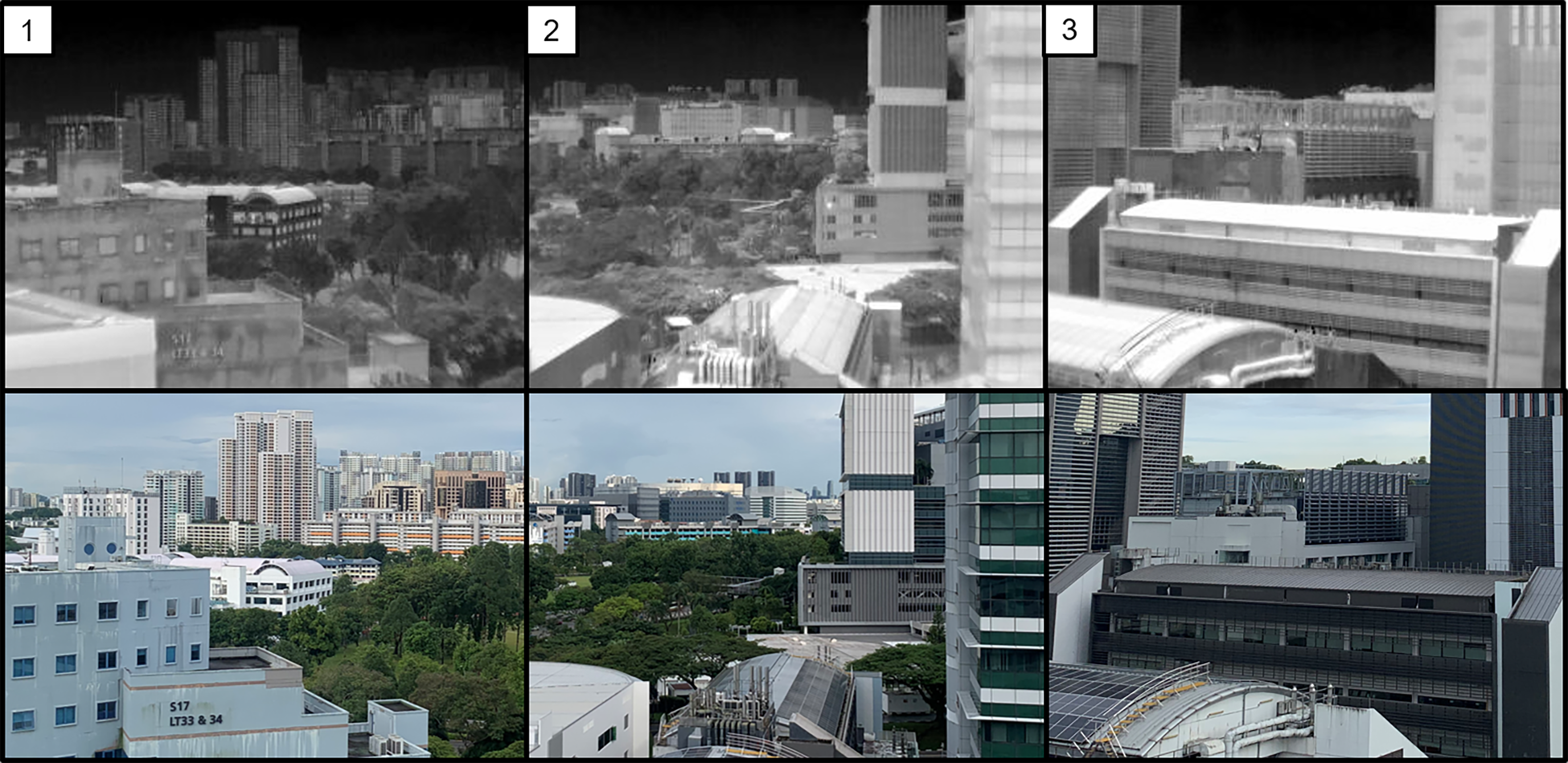}
\caption{Positions where the observatory captured thermal images on S16}
\label{fig:position_123}
\end{figure*}

The S16 observatory is located on the rooftop of a nine-story building in an educational area on a university campus. The pan/tilt unit was configured to capture images from three distinct targets, denoted as Positions 1, 2, and 3 in Figure \ref{fig:position_obser_s16}, sequentially at different time intervals. The observatory can capture information from three different buildings with thermal images (Figure \ref{fig:position_123}). The observatory moved from Position 1 to 3 sequentially and recorded thermal images at various time intervals per view. The thermal images obtained in this study elucidate a range of elements, including air conditioning units, glass materials, vegetation, and solar panels. These elements can be systematically analyzed to attain an in-depth understanding of their thermal behavior and consequential impact on the ambient environment. Specifically, this research is poised to investigate the influence of these elements on local temperature variations, energy consumption patterns, and the phenomena of urban heat islands. These insights carry significant implications for urban planning and the development of efficient energy management strategies.

\begin{figure*}[h!]
\centering
\includegraphics[width=\linewidth]{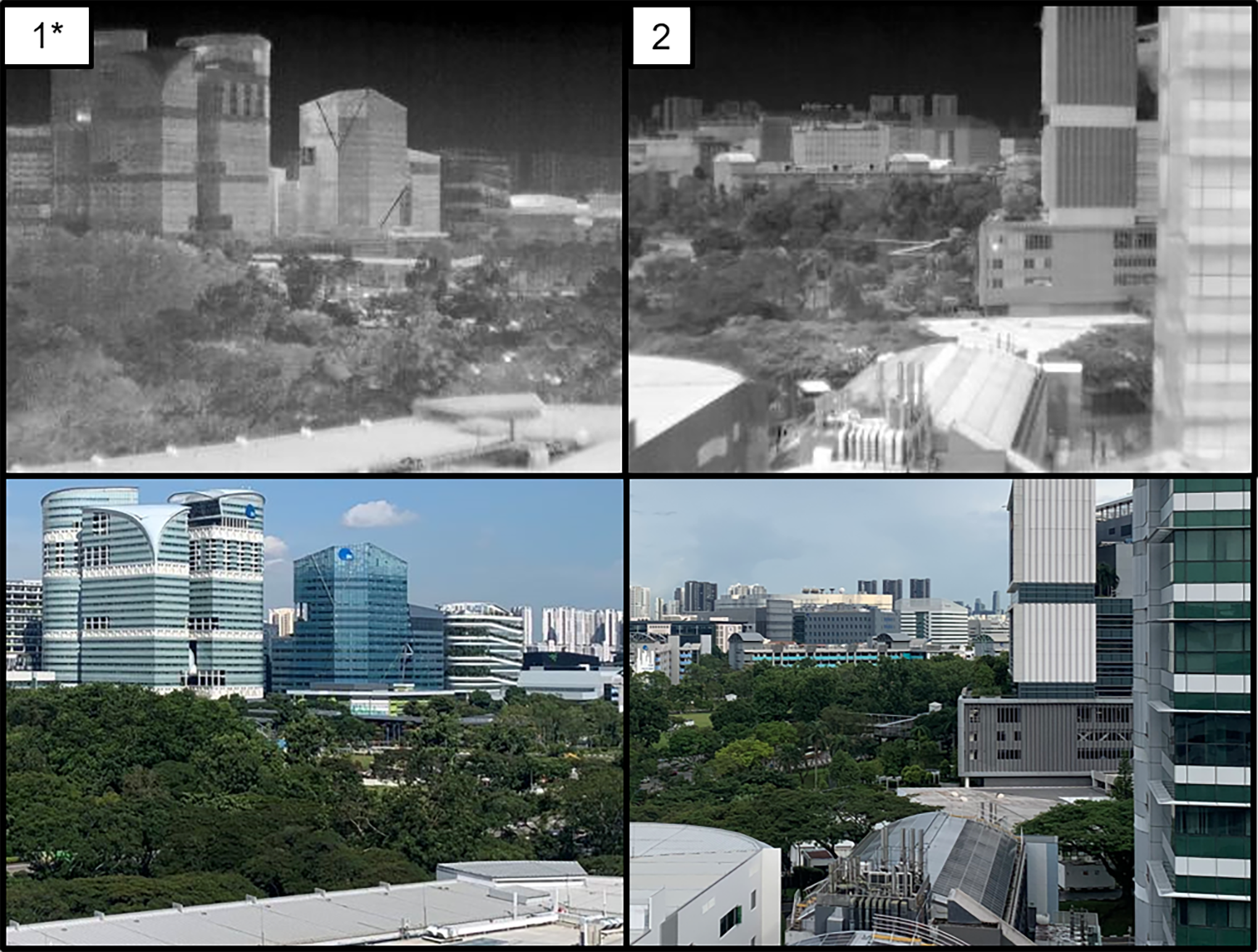}
\caption{Positions where the observatory captured thermal images on S16 after adjustment}
\label{fig:position_1*23}
\end{figure*}

Due to the necessary adjustments to the second observatory's positioning system, thermal images at the S16 site were not consistently captured from the exact same locations—slight variations to the left or right were noted. Consequently, automated segmentation tools are strongly recommended to account for these minor positional differences. It is acknowledged that these variations may impact the interpretation of temporal patterns in the data. Future enhancements to the observatory system are intended to improve the stability of image capture, leading to more consistent data collection. Furthermore, the positions from which images were captured were altered due to network issues with the observatory system on both Kent Value and S16. In an effort to achieve greater stability in the image capture locations on S16, measures were taken to ensure these positions were closely aligned with each other. Figure \ref{fig:position_1*23} depicts this more stable configuration.

The thermal camera comprises an optical system that directs radiation from the scene onto a microbolometer \cite{https://doi.org/10.48550/arxiv.2211.09288}. Long-wave infrared radiation changes the resistance of the detector, which is translated to apparent temperature ($T_{obj}$) readings as follows \cite{https://doi.org/10.48550/arxiv.2211.09288}:

\begin{equation}
\label{eq:T_obj}
T_{obj} = \frac{B}{\ln(\frac{R_1}{R_2(U_{obj}+O)}+f)}
\end{equation}

where $U_{obj}$ is the total signal response to the incident long-wave infrared radiation on the detector. This response is influenced by the material's emissivity, denoted by $\epsilon$, and the current environmental conditions. For a comprehensive computation of the influence factors for $U_{obj}$, please refer to Equation \ref{eq:U_all}. The other terms, $B$, $R_1$, $R_2$, $O$, and $f$, are calibration constants determined by the camera manufacturer in a controlled environment and are stored in the thermal image metadata. However, these parameters usually need to be re-calibrated in an outdoor environment where a significant discrepancy can be observed between $U_{obj}$ and $T_{obj}$. The re-calibrated parameters are presented in Table \ref{tab:calibration}.\\
Emissivity ($\epsilon$) measures a material's ability to emit infrared energy. Different materials have different emissivities, which can significantly impact the surface temperature measurements. Therefore, accounting for the material's emissivity under observation is crucial when processing thermal images. The dataset is presented in a radiometric format to preserve the raw thermal data, facilitating necessary corrections based on specific emissivity values. The total signal response $U_{obj}$ is also affected by ambient conditions, such as air temperature and humidity. By incorporating local weather data into the calculations, these influences can be corrected, and the accuracy of the surface temperature measurements can be improved. Please refer to the Usage Notes section for further information.\\

\subsection*{Data preprocessing}
Data preprocessing was done before analyzing the images for the Kent Vale dataset. First, data filtering was conducted to manually remove images unsuitable for analysis. Thermal images captured during the rainy period were disturbed by rain droplets and needed removal. Then, a convolution neural network (CNN) classification model was applied to exclude blurred images taken while the pan/tilt unit was moving. The CNN model is also used to classify the building type for the first location. The model was trained on 4316 pre-classified images and then used to classify thermal image datasets of Kent Vale.

\subsection*{Weather station data collection}
\begin{figure*}[h!]
\centering
\includegraphics[width=\linewidth]{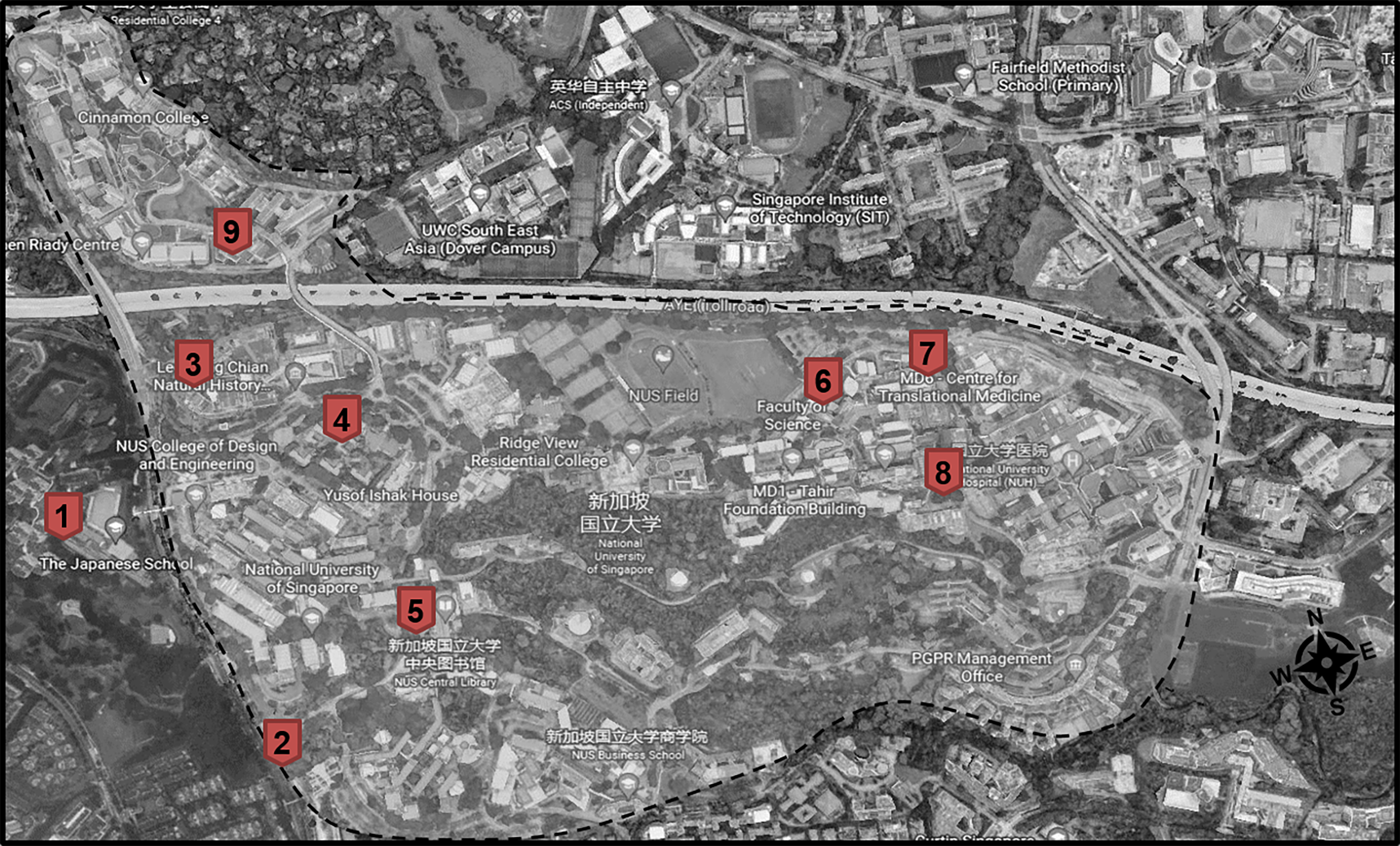}
\caption{Weather station locations \cite{MSS:Googleearth:Singapore, yu2020dependence, chen2022atlas}}
\label{fig:weather}
\end{figure*}

The weather station locations are shown in Figure \ref{fig:weather}. The weather stations measured meteorological data in multiple locations on the university campus around 2 meters above the ground, except for Locations 1 and 7 on rooftops. The six recorded parameters are air temperature (\degree C), relative humidity (RH, \%), dew point (\degree C), wind speed (m/s) and direction (degree), gust speed (m/s), and solar radiation (W/m$^2$) at a 1-minute interval. For location 9, only the first three parameters were recorded. The weather stations were calibrated before installation. The weather station specifications are listed in Table \ref{tab:ws}. The weather station installment details are available in Chen et al.\cite{chen2022atlas}, and Yu et al.\cite{yu2020dependence}.

\begin{table*}[h!]
\centering
\begin{tabular}{|l|l|l|l|l|}
\hline
Duration & Parameter & Temperature & Solar Radiation & Wind Speed \\
\hline 
\multirow{3}{*}{ \makecell{01/11/2021-08/03/2022 \\ 01/08/2022-14/12/2022}} & Range & -40\degree C - 75\degree C & 0 - 1280 $W/m^{2}$ & 0-50 m/s\\
\cline{2-5}
& Accuracy & ± 0.21 \degree C & ±10 $W/m^{2}$ & 0.2 m/s \\
\cline{2-5}
& Resolution & 0.01 \degree C & 1.25 $W/m^{2}$ & 0.2 m/s \\
\hline
\end{tabular}
\caption{\label{tab:ws} Specifications of weather station sensors \cite{yu2020dependence}}
\end{table*}

\section*{Data Records}
\label{sec:data_records}

\begin{figure*}[h!]
\centering
\includegraphics[width=\linewidth]{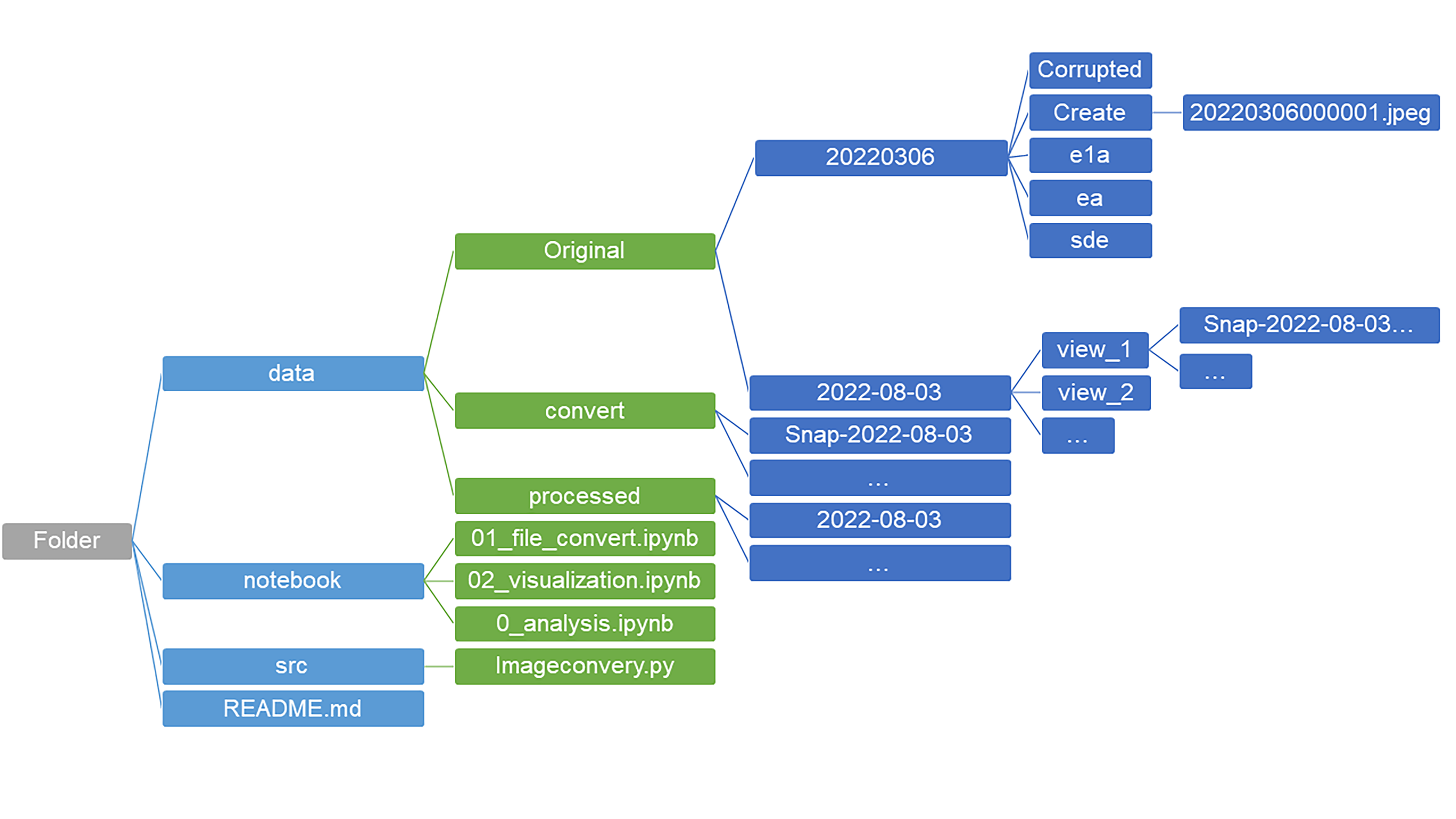}
\caption{File structure demonstration example for the dataset}
\label{fig:file_structure}
\end{figure*}

The thermal image and weather station data for the same periods are published in Zenodo with open access \cite{IRIS:thermal_image:2022}. The thermal images dataset contains two locations: Kent Vale and S16. A set of 483,915 thermal images were taken from Kent Vale between November 8, 2021, and March 8, 2022, while 882,006 thermal images were taken from S16 between August 3, 2022, and December 14, 2022. 
The images taken time are indicated in their file names. The data were converted and stored in JPEG format. The thermal images were stored in folders according to locations and views and listed sequentially. The detailed hierarchy of the data folders is listed in the data/README file on GitHub (\url{https://github.com/buds-lab/project-iris-dataset}) and Figure \ref{fig:file_structure}. Missing values are intended, as the preprocessing process filters unsuitable images for analysis. The CNN model classified the images taken at Kent Vale into five different groups: CREATE (I), E1A (II), EA (III), SDE4 (IV), and Corrupted. The Corrupted group collected images that were not classified into any of the above four views, which means that these collected images were of low quality and unable to be distinguished; hence, they needed to be excluded. The images taken at S16 were not entirely classified; while some were divided into three different viewpoints (view\_1, view\_2, and view\_3), some were not, requiring further classification. The summary of the provided dataset is shown in Table \ref{tab:summary_data}. In Zenodo, the Kent Vale dataset is stored in four independent zip files in sequential order, while the S16 dataset is stored in six independent zip files in sequential order. In each subcategory, the data is zipped in days. The code structure demonstration is also available in the Readme file in the GitHub repository. Examples of the data folder hierarchy include: 2022-08-03/view\_1/smap-2022-08-03T13-48-18.19.jpeg; 20220306/create/20220306000001.jpeg. The code demonstration contains three subcategories: data, notebook, and src. The data directory shows all the original and converted/processed data; The notebook directory includes three notebooks containing three preprocess steps; and the src directory runs preprocessing scripts. A more detailed description of the code file organization and explanation is available in the Code Availability section.  
The weather station data is in Excel format and stored in Zenodo as WS\_data.zip. It contains some missing values; therefore, prior data preprocessing is needed.

\begin{table*}[h!]
\centering
\begin{tabular}{|l|l|l|l|l|l|l|}
\hline
Duration & Resolution (sec) & Locations & Views & Examples & Corrupted size & Dataset size\\
\hline
\multirow{4}{*}{08/11/2021-08/03/2022} & \multirow{4}{*}{vary} & \multirow{4}{*}{Kent Vale} & I & 72036 & \multirow{4}{*}{100758} & \multirow{4}{*}{483915}\\
\cline{4-5}
& & & II & 96289 &  & \\
\cline{4-5}
& & & III & 127506 &  & \\
\cline{4-5}
& & & IV & 87326 &  & \\
\hline
\multirow{1}{*}{03/08/2022-14/12/2022} & \multirow{1}{*}{vary} & \multirow{1}{*}{S16} & no classify  & / & / & 882006\\
\hline
\end{tabular}
\caption{\label{tab:summary_data} Summary of the provided datasets at each location: duration of each location, number of views per group, the number of examples per view, corrupted size per view and dataset size}
\end{table*}

\section*{Technical Validation}
\label{sec:technical_validation}

\subsection*{Sensitivity analysis}
Before evaluating the infrared receptor ($T_{ij}$) from the observatory, a sensitivity analysis on factors that might significantly impact its variance was carried out. First-order Sobol index \cite{sobol2001global} was used to estimate the contribution of a variable to the variance of $T_{ij}$. The total Sobol index was also computed to consider the variables' interactions.

During the sensitivity analysis, the variables and their corresponding constraints are measured and shown in Table \ref{tab:constrains} \cite{https://doi.org/10.48550/arxiv.2210.11663}. The Satteli sampler \cite{saltelli2010variance} on thermal images taken at the four positions during a sunny day in Singapore was used to calculate the first order and total Sobol indices. $T_{ij}$ was determined for each sample at various times of the day and thermal camera placements. The sensitivity of each variable at different periods was given using the mean indices across the four positions.

\begin{table*}[h!]
\centering
\begin{threeparttable}
\begin{tabular}{|l|l|l|}
\hline
Variable & Minimum value & Maximum value \\
\hline
Thermal emissivity (-) & 0.7 & 1.0 \\
\hline
Sky temperature (\degree C)\tnote{*} & 11.0 & 33.0 \\
\hline
Air temperature (\degree C)\tnote{**} & 19.0 & 37.0 \\
\hline
Air relative humidity (\%)\tnote{**} & 30.0 & 100.0 \\
\hline
Distance object (m) & 100.0 & 1000.0 \\
\hline
Window temperature (\degree C)\tnote{+} & 20.0 & 60.0 \\
\hline
Window transmittance (-) & 0.8 & 1.0 \\
\hline
\end{tabular}
\begin{tablenotes}
\footnotesize
\item[*] The sky temperature ($T^{sky} = g(U^{sky})$) was measured by Miguel et al. \cite{https://doi.org/10.48550/arxiv.2210.11663}
\item[**] Weather data recorded by the Meteorological Service of Singapore \cite{MSS:hisdaily:2022}
\item[+]Measured from a HOBO UX100-014M sensor
\end{tablenotes}
\caption{\label{tab:constrains} Constrains of each variable considered during the sensitivity analysis of $T_{ij}$}
\end{threeparttable}
\end{table*}

\subsection*{Parameters calibration}
\begin{figure*}[h!]
\centering
\includegraphics[width=\linewidth]{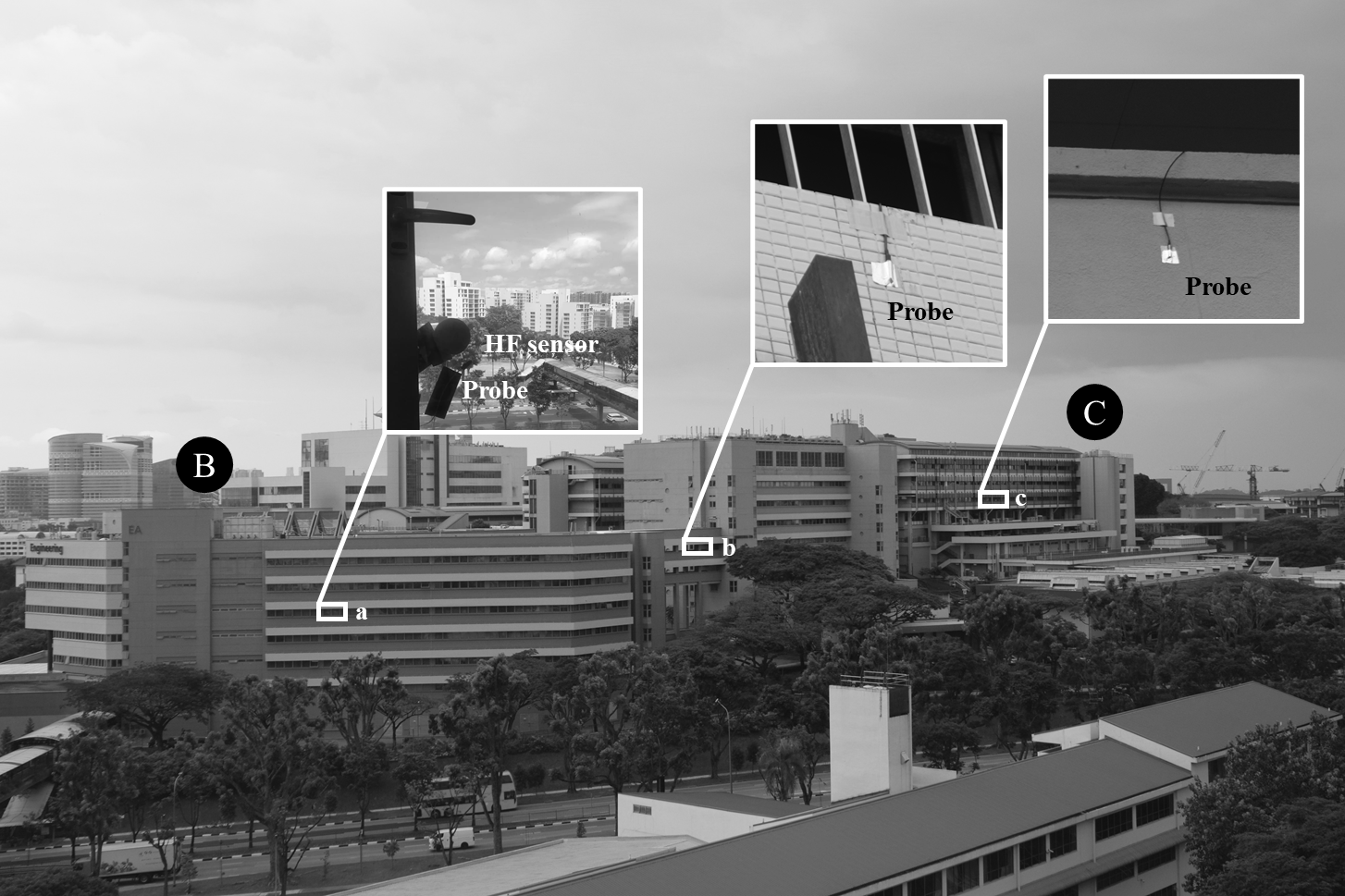}
\caption{Locations where sensors were placed in to calibrate thermal images and measure the surface temperature \cite{https://doi.org/10.48550/arxiv.2210.11663}}
\label{fig:calibration}
\end{figure*}

The thermal images' surface temperature estimates were calibrated in accordance with the information gathered by contact surface sensors, as shown in Figure \ref{fig:calibration} at three different positions. Between December 2021 and March 2022, a heat flux (HF) sensor and a temperature probe were installed at Position $a$ on the surface of a window in Building B. The HF sensor and the temperature probe were connected to a Hioki data logger and measured the surface temperature every 3 seconds. Another type of data logger, the UbiBot WS1 pro indoor monitoring sensor, was used to connect temperature probes at position $b$ and $c$, respectively. 

The output voltage recorded by the infrared receptor at position $ij$ in the thermal image is represented by an array $U^{r}_{ij}$ and a header in these files. Equation \ref{eq:U} can be used to transform $U^{r}_{ij}$ into longwave radiations ($L^{r}_{ij}$) between 7.5 and 13 micrometers:

\begin{equation}
\label{eq:U}
U^{r}_{ij} = cL^{r}_{ij}
\end{equation}
where the term $c$ is not truly constant but depends on the emissivity of the surface material being observed, with lower emissivities reflecting more radiation from other sources. 

The FLIR A300 (9 Hz) thermal camera that was installed in the observatory was calibrated such that the following Equation \ref{eq:T} holds true for the surface temperature of a target element detected by the infrared receptor ($T^{r}_{ij}$):

\begin{equation}
\label{eq:T}
T^{r}_{ij} = 
g(U^{r}_{ij}) =
b\ln \left[ \frac{r_{1}}{r_{2}(U_{ij}+O)}+f \right] ^{-1}
\end{equation}
where $b$, $r_{1}$, $r_{2}$, $O$, and $f$ are calibrated parameters.

However, the parameters in Equation \ref{eq:T} need to be calibrated in order to minimize the discrepancy between $T^{r}_{ij}$ and $T_{ij}$ in an outdoor environment. In this case, these values are acquired in a controlled setting where the target element's actual surface temperature $T_{ij}$ at location $ij$ in the thermal picture is almost identical to $T^{r}_{ij}$.

\begin{figure*}[h!]
\centering
\includegraphics[width=\linewidth]{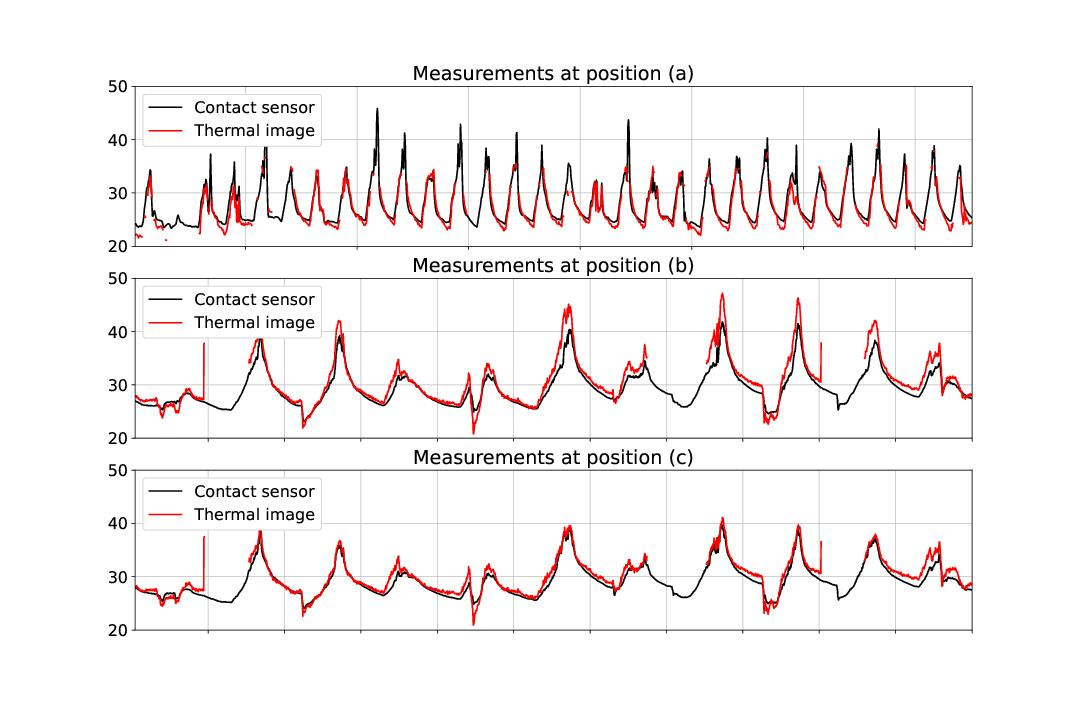}
\caption{Comparison of estimated and measured surface temperatures after calibration of the A300 infrared camera at the observatory. \cite{https://doi.org/10.48550/arxiv.2210.11663}}
\label{fig:comparison}
\end{figure*}

The FLIR A300 camera was calibrated by manually tuning the parameters in Equation \ref{eq:T} until there was a satisfactory agreement between surface temperature estimations and observations. The agreement was assessed in terms of the Mean Bias Error (MBE) and Root Mean Square Error (RMSE) as shown in Equation \ref{eq:mbe} and \ref{eq:rmse}:

\begin{equation}
\label{eq:mbe}
MBE = \frac{1}{n}\Sigma_{i=1}^{n}{\Big(T^{n}_S -T^{n}_{S,m}\Big)^2}
\end{equation}

\begin{equation}
\label{eq:rmse}
RMSE = \sqrt{\frac{1}{n}\Sigma_{i=1}^{n}{\Big({T^{n}_S -T^{n}_{S,m}}\Big)^2}}
\end{equation}

where $T^{n}_{S,m}$ is the surface temperature obtained by contact surface sensors, and $T^{n}_S$ is the surface temperature calculated from thermal pictures captured by the FLIR A300 camera at time $t = t_0 + n\cdot \Delta t$. The calibration aims to achieve an MBE as close to 0 with the lowest RMSE. The RMSE and MBE were calculated with different $\Delta t$ at three different positions to optimize the temporal resolution for each location (30 minutes at Position A and 5 minutes for Positions B and C). At Positions A-C, the RMSE and MBE obtained after calibration are shown in Table \ref{tab:calibration}, achieving an RMSE below 2 degrees Celsius and an MBE below ± 1 degree Celsius. Furthermore, Figure \ref{fig:comparison} illustrates a comparison between the estimated and measured surface temperatures after calibration at Positions A-C, offering additional information that enables the data to be freely accessed and tailored to suit specific needs and interests in further research.

\begin{table}[ht]
\centering
\begin{threeparttable}
\begin{tabular}{|l|l|l|}
\hline
Constant & Original parameters & Calibrated parameters \\
\hline
B & 1396.6000 & 1425.0000 \\
\hline
$f$ &1.0000 & 1.0000 \\
\hline
O & -6303.0000 & -9550.0000 \\
\hline
R1 & 14911.1846 & 14911.1850 \\
\hline
R2 & 0.0108 & 0.0120 \\
\hline
\hline
Position & RMSE (in \degree C) & MBE (in \degree C) \\
\hline
A & 1.24 & -0.21 \\
\hline
B & 1.69 & 0.91 \\
\hline
C & 1.00 & 0.55 \\
\hline
\end{tabular}
\caption{\label{tab:calibration} Calibration parameters of the measurement\cite{https://doi.org/10.48550/arxiv.2210.11663}.}
\end{threeparttable}
\end{table}

\section*{Usage Notes}
\label{sec:usage_notes}

\subsection*{Factors consideration}
When evaluating $T_{ij}$ from the rooftop observatory, several factors must be considered. First, the buildings, streets, and trees that could be seen from the observatory are exposed to longwave radiations mainly coming from the skydome. Then, in tandem with the longwave radiation released by the element ($L$) being observed from the observatory, the longwave radiation from the sky ($L^{sky}$) is reflected into the air. Both $L^{sky}$ and $L$ pass through the atmosphere before arriving at the observatory, then together with the longwave radiation emitted by the atmosphere ($L^{atm}$), transmitting through the window of the housing before reaching the infrared receptor. Along with other types of radiation, the infrared receptor also captures longwave radiation from the window ($L^{win}$). The camera output voltage corresponding to $U_{ij}$ combining the relationship defined in Equation \ref{eq:T} and the aforementioned factors could be expressed as the following equation :

\begin{equation}
\label{eq:U_all}
U_{ij} 
= \frac{1}{\varepsilon_{ij}\tau^{atm}_{ij}\tau^{win}}U^{r}_{ij}
- \frac{1-\varepsilon_{ij}}{\varepsilon_{ij}}U^{sky}
- \frac{1-\tau^{atm}_{ij}}{\varepsilon_{ij}\tau^{atm}_{ij}}U^{atm}
- \frac{1-\tau^{win}}{\varepsilon_{ij}\tau^{atm}_{ij}\tau^{win}}U^{win}
\end{equation}

where $\varepsilon_{ij}$ is the thermal emissivity from the element at position $ij$ in the thermal image, $\tau^{atm}_{ij}$ is the transmissivity of the atmosphere between the element and the rooftop observatory, and $\tau^{win}$ is the transmissivity of the window.

$T_{ij}$ could be calculated through Equation \ref{eq:T} if $U_{ij}$ is known. $\varepsilon_{ij}$ and $\tau^{win}$ could be estimated from the material properties of the element and the window, respectively. $\tau^{atm}_{ij}$ can be calculated from weather data collected from a weather station.

Considering building material properties is paramount in analyzing heat fluxes, as various materials exhibit distinct thermal behaviors. Specifically, Building A's facade comprises two steel walls and one glass wall. In contrast, Buildings B and C predominantly feature concrete walls, while Building D's facade uniquely incorporates steel and concrete walls. The diversity in materials necessitates a detailed understanding of their emissive properties, which are critical in accurately calculating temperatures. Any oversight in this regard could potentially lead to misleading results. Therefore, the thermal properties, including the corresponding emissivities of the building facades and vegetation, have been meticulously defined. The thermal properties of the building facades and vegetation, including the corresponding emissivities, are defined and presented in Table \ref{tab:thermal_properties}, which has been constructed based on the relevant literature.

\begin{table*}[h!]
\centering
\begin{tabular}{|l|l|l|l|l|l|l|}
\hline
\thead{Portion \\ } & \thead{Emissivity \\(0-1)} & \thead{Density \\ ($kg/m^3$)} & \thead{Specific heat \\(J/kg-K)}  & \thead{Thickness \\(cm)} & \thead{Thermal conductivity \\ ((W/m-K))} & \thead{Reference\\} \\
\hline 
Steel & 0.88 & 7940 & 507 & 10 & - & \cite{miguel2021physically, meneghetti2013synthesis}\\
Glass & 0.93 & 2500 & 836 & 2.5 & 0.974 & \cite{raman1982thermal, ritland1954density, sharp1951effect, wang2020optically} \\
Concrete & 0.90 & 2400 & 1180 & 30 & 0.62 & \cite{miguel2021physically, iffat2015relation, de1995specific, kim2003experimental}\\
\hline
\thead{Portion \\ } & \thead{Emissivity \\(0-1)} & \thead{leaf area index\\ ($m^2/m^2$)} &&&& \thead{Reference\\}\\
\hline
\thead{Vegetation \\ (Tropical trees) }&0.98 & 4  &&&&\cite{olioso1995simulating, ganguly2008generating} \\
\hline
\end{tabular}
\caption{\label{tab:thermal_properties} Thermal properties of building facades and vegetation \cite{https://doi.org/10.48550/arxiv.2210.11663}}
\end{table*}

\subsection*{Weather stations data}
The weather information obtained from the weather stations could be used together with the thermal images gathered by the rooftop observatory to estimate urban heat fluxes. Eight weather stations measured air temperature, relative humidity, dew point, wind speed and direction, gust speed, and solar radiation.

Weather stations near the observatory camera could be used to calculate $\tau^{atm}_{ij}$ in the following equation according to Waldemar and Klecha \cite{waldemar2015modeling,https://doi.org/10.48550/arxiv.2210.11663}:
\begin{equation}
\label{eq:tau_ij}
\tau^{atm}_{ij}
= x \cdot \exp\lbrack -\sqrt{d_{ij}}(\alpha_1 + \beta_1 \sqrt{\omega})\rbrack + (1-x) \cdot \exp\lbrack-\sqrt{d_{ij}}(\alpha_2 + \beta_2 \sqrt{\omega})\rbrack
\end{equation}

where $d_{ij}$ is the distance between the element and the observatory at position $ij$ of the thermal image, $\omega$ is the atmosphere's water vapor content, $\alpha$, $\beta$, and $x$ are empirical coefficients. $\omega$ can be estimated from the air temperature ($T_{air}$) and relative humidity ($\phi$) measured from the weather stations using the following equation:

\begin{equation}
\label{eq:omega}
\omega
= \phi \cdot \exp\lbrack \gamma_o + \gamma_1 T_{air} + \gamma_2 T^{2}_{air} + \gamma_3 T^{3}_{air} \rbrack
\end{equation}

where $\gamma$ is another set of empirical coefficients described in Waldemar and Klecha \cite{waldemar2015modeling,https://doi.org/10.48550/arxiv.2210.11663}

\subsection*{Preprocessing}
Following the classification of CNN, segmentation of the desired region and extraction of its radiometric data are required in order to evaluate the dataset. Flirextractor python package \cite{MSS:flirextractor:2020} is recommended, as it allows extracting temperature data from regions of interest in the thermal images and storing it in CSV (comma-separated values) format with a simple process. The temperature data from thermal images were converted using Equation \ref{eq:T_obj}. The python package Labelme \cite{russell2008labelme} is helpful in the segmentation of regions of interest for analysis. The relevant codes are provided in the Code Availability section.

\subsection*{Image information}
The thermal image data are classified and stored based on their positions. The image datasets contain information including time, device positioning, and observatory locations (Kent Vale or S16). The building detail and surrounding environment could be seen based on the corresponding real images in this work. Analysis can be conducted in the time domain, frequency domain, or both, based on the nature of the time series and the degree of information to be retrieved \cite{https://doi.org/10.48550/arxiv.2211.09288}.

\subsection*{Potential applications}
The thermal images provide surface temperatures of urban elements such as buildings, roads, and vegetation. Some example applications of surface temperature data are urban heat island analysis \cite{https://doi.org/10.48550/arxiv.2210.11663,weng2018urban}, urban energy monitoring \cite{yang2021urban}, and building thermal performance monitoring \cite{balaras2002infrared, kylili2014infrared, https://doi.org/10.48550/arxiv.2211.09288}. A detailed application of urban observatory is available in Dobler's work \cite{dobler2021urban}.
Thermal images collected by the observatory can provide high-resolution data to study dynamic interactions in buildings at the district scale, much smaller than the city scale captured by satellite thermal image data, which are used frequently in relevant fields. 
Further studies of the thermal properties of non-residential buildings can be conducted and compared with those estimated at the city scale.
The image data also lends itself to several digital image processing techniques, including edge detection, image enhancement, segmentation, and statistical analysis. For instance, edge detection could help discern different elements within the urban environment, such as buildings, vegetation, and transportation infrastructure. Image enhancement techniques could enhance the visibility of thermal patterns, and segmentation could be used to isolate areas of interest. The dataset provided here is well-suited for various applications, especially those involving urban heat island effects, building energy efficiency studies, and urban microclimate analysis. However, there remains potential for improving the accuracy of surface temperature measurements. Discussions about the data and innovative approaches to its interpretation and application are welcomed.

\subsection*{Privacy and safety control}
These thermal images form the foundation for various research analyses related to urban living environments. For instance, they can be used to study temperature variations within an urban block, understand the effect of building materials on heat retention, or analyze the influence of vegetation on local microclimates. Researchers are encouraged to use various analytical techniques to probe the data, including computational statistics, machine learning, and sequential analysis. Machine learning, for instance, could be used to predict future temperature patterns based on past data, while sequential analysis could help understand temporal changes in the thermal landscape. However, data misuse needs to be prevented \cite{dobler2021urban,floridi2018ai4people}. Since IoT cameras can be easily abused \cite{tom2017designing}, the dataset collected here should include sufficient privacy protections for relevant individuals and cities \cite{dobler2021urban,lane_stodden_bender_nissenbaum_2014}. In such case, the images collected in this work do not contain personally identifiable information \cite{mccallister2010guide} such as facial features, and the individual feature cannot be tracked. All taken images were strictly limited in pixel resolution so that interiors of buildings cannot be seen \cite{dobler2021urban}.

\section*{Code Availability}
\label{sec:code_availability}

Python 3 was used for data processing and analysis. The simple code demonstration suggested in this work can be found in the GitHub repository stated in the Data Record section. The code demonstrates the methodology for extraction and subsequent processing of the thermal images. Few images were selected from the dataset for illustration purposes.
The code uses two specific GitHub packages: Flirextracter \cite{MSS:flirextractor:2020} and Labelme \cite{russell2008labelme}. Flirextractor is an efficient Python package for extracting temperature data from thermal images and converting it into an array, then saving it as a CSV file for further access. The link to the GitHub package describing the usage of Flirextractor in detail is as follows: \url{https://github.com/aloisklink/flirextractor}. Labelme is a graphical image annotation tool written in Python. It allows users to demarcate the desired area of any shape with simple mouse clicks. Detailed instructions for Labelme can be found at the following link: \url{https://github.com/wkentaro/labelme}. 
The GitHub code repository exhibits A meticulously structured hierarchy designed to enhance navigation and understandability. It incorporates three primary directories: 'data', 'notebook', and 'src'. The 'data' directory segregates original, converted, and processed files, whereas the 'notebook' directory encompasses Jupyter notebooks detailing file conversion, data visualization, and data analysis procedures. The 'src' directory hosts a Python script dedicated to image conversion.
Furthermore, the 'data' folder manifests a systematic organization structured to efficiently manage different stages of data processing. It consists of three primary subdirectories: 'original', 'convert', and 'labelme'. The ‘original’ and ‘processed’ subdirectories mirror each other, containing dated folders representing different acquisition dates, with each date folder further divided into ‘view\_1’, ‘view\_2’, and ‘view\_3’ subdirectories. These subdirectories contain the respective images captured from each view named “smap-YYYY-MM-DDTHH-MM-SS.MS.jpeg”. During the data processing phase, corresponding 'json' files for each image are generated in the 'labelme' directory. In addition, for each image, a dedicated directory is established within the 'convert' subdirectory, encapsulating original, processed, and labeled images and a text file enumerating the detected labels.
Detailed file structure tree plots, included within the respective README files, furnish comprehensive visual representations of the file and folder organization, thus facilitating an understanding of the hierarchical structure and interrelationships among different components of the codebase. Moreover, an example involving a representative image has been conducted to illustrate the efficacy of the employed image processing techniques; the results highlight the segmentation, further processing, and final presentation of the image augmented with enhanced colors and informative labels. Please contact the corresponding author directly if you have any particular requirements.


\section*{Acknowledgements}

This research is supported by the National Research Foundation and the Prime Minister’s Office of Singapore under its
Campus for Research Excellence and Technological Enterprise
(CREATE) program. It was funded through a grant to the Berkeley Education Alliance for Research in Singapore (BEARS) for
the Singapore-Berkeley Building Efficiency and Sustainability in
the Tropics (SinBerBEST2) program. BEARS was established by
the University of California, Berkeley, as a center for intellectual excellence in research and education in Singapore. 
This research is also supported by the Multi-scale Digital Twins for the Urban Environment: From Heartbeats to Cities project, funded by the Singapore Ministry of Education (MOE) Tier 1 Academic Research Fund, NUS Resilience and Growth Postdoctoral Fellowship - Smart Cities and Urban Analytics supported by National Research Foundation (NRF), and the Johnson Controls Open Blue Innovation Laboratory. Furthermore, we would like to acknowledge the contributions of the Development of Integrated Multi-scale and Multiphysics Urban Microclimate Model project, funded by the National University of Singapore and supported by the University Campus Infrastructure (UCI) and Office of the Deputy President (Research \& Technology).

\section*{Author contributions statement}
S.L. wrote the original draft and assisted in data collection and formal analysis, V.S. and M.M. designed and deployed the data collection platform, P.A. and M.I. supported the data collection process, A.C. and F.B. provided conceptual input for the data platform deployment and reviewed the manuscript, K.P. edited and reviewed the manuscript. and C.M. supervised the deployment, data collection, and development outlined in this article and edited and reviewed the manuscript. All authors have reviewed the final manuscript.

\section*{Competing interests} 

The authors declare no competing interests.

\bibliographystyle{model1-num-names}
\bibliography{00_main}

\begin{thebibliography}{43}
\expandafter\ifx\csname natexlab\endcsname\relax\def\natexlab#1{#1}\fi
\providecommand{\bibinfo}[2]{#2}
\ifx\xfnm\relax \def\xfnm[#1]{\unskip,\space#1}\fi
\bibitem[{Dobler et~al.(2021)Dobler, Bianco, Sharma, Karpf, Baur, Ghandehari, Wurtele, and Koonin}]{dobler2021urban}
\bibinfo{author}{G.~Dobler}, \bibinfo{author}{F.~B. Bianco}, \bibinfo{author}{M.~S. Sharma}, \bibinfo{author}{A.~Karpf}, \bibinfo{author}{J.~Baur}, \bibinfo{author}{M.~Ghandehari}, \bibinfo{author}{J.~Wurtele}, \bibinfo{author}{S.~E. Koonin},
\newblock \bibinfo{title}{The urban observatory: a multi-modal imaging platform for the study of dynamics in complex urban systems},
\newblock \bibinfo{journal}{Remote Sensing} \bibinfo{volume}{13} (\bibinfo{year}{2021}) \bibinfo{pages}{1426}.
\bibitem[{Liu and Biljecki(2022)}]{2022_jag_geoai}
\bibinfo{author}{P.~Liu}, \bibinfo{author}{F.~Biljecki},
\newblock \bibinfo{title}{A review of spatially-explicit geoai applications in urban geography},
\newblock \bibinfo{journal}{International Journal of Applied Earth Observation and Geoinformation} \bibinfo{volume}{112} (\bibinfo{year}{2022}) \bibinfo{pages}{102936}.
\bibitem[{MSS(2022)}]{MSS:ieabuilding:2022}
\bibinfo{title}{Iea (2022), buildings, tracking report}, \bibinfo{howpublished}{\emph{buildings} \url{https://www.iea.org/reports/buildings}}, \bibinfo{year}{2022}.
\bibitem[{Miguel et~al.(2021)Miguel, Hien, Marcel, Chung, Yueer, Zhonqi, Ji-Yu, Raghavan, and Son}]{miguel2021physically}
\bibinfo{author}{M.~Miguel}, \bibinfo{author}{W.~N. Hien}, \bibinfo{author}{I.~Marcel}, \bibinfo{author}{H.~D.~J. Chung}, \bibinfo{author}{H.~Yueer}, \bibinfo{author}{Y.~Zhonqi}, \bibinfo{author}{D.~Ji-Yu}, \bibinfo{author}{S.~V. Raghavan}, \bibinfo{author}{N.~N. Son},
\newblock \bibinfo{title}{A physically-based model of interactions between a building and its outdoor conditions at the urban microscale},
\newblock \bibinfo{journal}{Energy and Buildings} \bibinfo{volume}{237} (\bibinfo{year}{2021}) \bibinfo{pages}{110788}.
\bibitem[{Lucchi(2018)}]{lucchi2018applications}
\bibinfo{author}{E.~Lucchi},
\newblock \bibinfo{title}{Applications of the infrared thermography in the energy audit of buildings: A review},
\newblock \bibinfo{journal}{Renewable and Sustainable Energy Reviews} \bibinfo{volume}{82} (\bibinfo{year}{2018}) \bibinfo{pages}{3077--3090}.
\bibitem[{Ngie et~al.(2014)Ngie, Abutaleb, Ahmed, Darwish, and Ahmed}]{ngie2014assessment}
\bibinfo{author}{A.~Ngie}, \bibinfo{author}{K.~Abutaleb}, \bibinfo{author}{F.~Ahmed}, \bibinfo{author}{A.~Darwish}, \bibinfo{author}{M.~Ahmed},
\newblock \bibinfo{title}{Assessment of urban heat island using satellite remotely sensed imagery: a review},
\newblock \bibinfo{journal}{South African Geographical Journal= Suid-Afrikaanse Geografiese Tydskrif} \bibinfo{volume}{96} (\bibinfo{year}{2014}) \bibinfo{pages}{198--214}.
\bibitem[{Almeida et~al.(2021)Almeida, Teodoro, and Gon{\c{c}}alves}]{almeida2021study}
\bibinfo{author}{C.~R.~d. Almeida}, \bibinfo{author}{A.~C. Teodoro}, \bibinfo{author}{A.~Gon{\c{c}}alves},
\newblock \bibinfo{title}{Study of the urban heat island (uhi) using remote sensing data/techniques: A systematic review},
\newblock \bibinfo{journal}{Environments} \bibinfo{volume}{8} (\bibinfo{year}{2021}) \bibinfo{pages}{105}.
\bibitem[{Balaras and Argiriou(2002)}]{balaras2002infrared}
\bibinfo{author}{C.~A. Balaras}, \bibinfo{author}{A.~Argiriou},
\newblock \bibinfo{title}{Infrared thermography for building diagnostics},
\newblock \bibinfo{journal}{Energy and buildings} \bibinfo{volume}{34} (\bibinfo{year}{2002}) \bibinfo{pages}{171--183}.
\bibitem[{Arjunan et~al.(2021)Arjunan, Dobler, Lee, Miller, Biljecki, and Poolla}]{2021_buildsys_acir}
\bibinfo{author}{P.~Arjunan}, \bibinfo{author}{G.~Dobler}, \bibinfo{author}{K.~Lee}, \bibinfo{author}{C.~Miller}, \bibinfo{author}{F.~Biljecki}, \bibinfo{author}{K.~Poolla},
\newblock \bibinfo{title}{{Operational characteristics of residential air conditioners with temporally granular remote thermographic imaging}},
\newblock in: \bibinfo{booktitle}{BuildSys '21: Proceedings of the 8th ACM International Conference on Systems for Energy-Efficient Buildings, Cities, and Transportation}, Proceedings of the 8th ACM International Conference on Systems for Energy-Efficient Buildings, Cities, and Transportation, pp. \bibinfo{pages}{184--187}.
\bibitem[{Sham et~al.(2012)Sham, Lo, and Memon}]{sham2012verification}
\bibinfo{author}{J.~F. Sham}, \bibinfo{author}{T.~Y. Lo}, \bibinfo{author}{S.~A. Memon},
\newblock \bibinfo{title}{Verification and application of continuous surface temperature monitoring technique for investigation of nocturnal sensible heat release characteristics by building fabrics},
\newblock \bibinfo{journal}{Energy and Buildings} \bibinfo{volume}{53} (\bibinfo{year}{2012}) \bibinfo{pages}{108--116}.
\bibitem[{Martin et~al.(2022)Martin, Chong, Biljecki, and Miller}]{martin2022infrared}
\bibinfo{author}{M.~Martin}, \bibinfo{author}{A.~Chong}, \bibinfo{author}{F.~Biljecki}, \bibinfo{author}{C.~Miller},
\newblock \bibinfo{title}{Infrared thermography in the built environment: A multi-scale review},
\newblock \bibinfo{journal}{Renewable and Sustainable Energy Reviews} \bibinfo{volume}{165} (\bibinfo{year}{2022}) \bibinfo{pages}{112540}.
\bibitem[{Chew et~al.(2021)Chew, Liu, Li, and Norford}]{chew2021interaction}
\bibinfo{author}{L.~W. Chew}, \bibinfo{author}{X.~Liu}, \bibinfo{author}{X.-X. Li}, \bibinfo{author}{L.~K. Norford},
\newblock \bibinfo{title}{Interaction between heat wave and urban heat island: A case study in a tropical coastal city, singapore},
\newblock \bibinfo{journal}{Atmospheric Research} \bibinfo{volume}{247} (\bibinfo{year}{2021}) \bibinfo{pages}{105134}.
\bibitem[{Roth(2007)}]{roth2007review}
\bibinfo{author}{M.~Roth},
\newblock \bibinfo{title}{Review of urban climate research in (sub) tropical regions},
\newblock \bibinfo{journal}{International Journal of Climatology: A Journal of the Royal Meteorological Society} \bibinfo{volume}{27} (\bibinfo{year}{2007}) \bibinfo{pages}{1859--1873}.
\bibitem[{Ignatius et~al.(2022)Ignatius, Xu, Hou, Liang, Zhao, Chen, Wong, and Biljecki}]{2022_3dgeoinfo_lcz_sg_svi}
\bibinfo{author}{M.~Ignatius}, \bibinfo{author}{R.~Xu}, \bibinfo{author}{Y.~Hou}, \bibinfo{author}{X.~Liang}, \bibinfo{author}{T.~Zhao}, \bibinfo{author}{S.~Chen}, \bibinfo{author}{N.~Wong}, \bibinfo{author}{F.~Biljecki},
\newblock \bibinfo{title}{Local climate zones: Lessons from singapore and potential improvement with street view imagery},
\newblock \bibinfo{journal}{ISPRS Annals of Photogrammetry, Remote Sensing and Spatial Information Sciences} \bibinfo{volume}{X-4/W2-2022} (\bibinfo{year}{2022}) \bibinfo{pages}{121--128}.
\bibitem[{Martin et~al.(2022)Martin, Ramani, and Miller}]{https://doi.org/10.48550/arxiv.2210.11663}
\bibinfo{author}{M.~Martin}, \bibinfo{author}{V.~Ramani}, \bibinfo{author}{C.~Miller}, \bibinfo{title}{Infrared investigation in singapore (iris) observatory: Urban heat island contributors and mitigators analysis using neighborhood-scale thermal imaging}, \bibinfo{year}{2022}.
\bibitem[{Ramani et~al.(2022)Ramani, Martin, Arjunan, Chong, Poolla, and Miller}]{https://doi.org/10.48550/arxiv.2211.09288}
\bibinfo{author}{V.~Ramani}, \bibinfo{author}{M.~Martin}, \bibinfo{author}{P.~Arjunan}, \bibinfo{author}{A.~Chong}, \bibinfo{author}{K.~Poolla}, \bibinfo{author}{C.~Miller}, \bibinfo{title}{Longitudinal thermal imaging for scalable non-residential hvac and occupant behaviour characterization}, \bibinfo{year}{2022}.
\bibitem[{Chen et~al.(2022)Chen, Wong, Ignatius, Zhang, He, Yu, and Hii}]{chen2022atlas}
\bibinfo{author}{S.~Chen}, \bibinfo{author}{N.~H. Wong}, \bibinfo{author}{M.~Ignatius}, \bibinfo{author}{W.~Zhang}, \bibinfo{author}{Y.~He}, \bibinfo{author}{Z.~Yu}, \bibinfo{author}{D.~J.~C. Hii},
\newblock \bibinfo{title}{Atlas: Software for analysing the relationship between urban microclimate and urban morphology in a tropical city},
\newblock \bibinfo{journal}{Building and Environment} \bibinfo{volume}{208} (\bibinfo{year}{2022}) \bibinfo{pages}{108591}.
\bibitem[{Yu et~al.(2020)Yu, Chen, Wong, Ignatius, Deng, He, and Hii}]{yu2020dependence}
\bibinfo{author}{Z.~Yu}, \bibinfo{author}{S.~Chen}, \bibinfo{author}{N.~H. Wong}, \bibinfo{author}{M.~Ignatius}, \bibinfo{author}{J.~Deng}, \bibinfo{author}{Y.~He}, \bibinfo{author}{D.~J.~C. Hii},
\newblock \bibinfo{title}{Dependence between urban morphology and outdoor air temperature: A tropical campus study using random forests algorithm},
\newblock \bibinfo{journal}{Sustainable Cities and Society} \bibinfo{volume}{61} (\bibinfo{year}{2020}) \bibinfo{pages}{102200}.
\bibitem[{MSS(2016)}]{MSS:Googleearth:Singapore}
\bibinfo{title}{1\degree17'50"n, 103\degree46'37"e. \textbf{Google Earth}. march 05, 2016. november 01, 2022}, \bibinfo{howpublished}{\url{https://earth.google.com/web/@1.29728934,103.77702845,23.33038212a,2430.67786248d,35y,27.55033278h,0t,0r}}, \bibinfo{year}{2016}.
\bibitem[{Lin et~al.(2022)Lin, Ramani, Martin, Arjunan, Chong, Biljecki, Ignatius, Poolla, and Miller}]{IRIS:thermal_image:2022}
\bibinfo{author}{S.~Lin}, \bibinfo{author}{V.~Ramani}, \bibinfo{author}{M.~Martin}, \bibinfo{author}{P.~Arjunan}, \bibinfo{author}{A.~Chong}, \bibinfo{author}{F.~Biljecki}, \bibinfo{author}{M.~Ignatius}, \bibinfo{author}{K.~Poolla}, \bibinfo{author}{C.~Miller}, \bibinfo{title}{Infrared investigation in singapore (iris) dataset}, \bibinfo{howpublished}{\emph{Zenodo} \url{https://zenodo.org/doi/10.5281/zenodo.7463995}}, \bibinfo{year}{2022}.
\bibitem[{Sobol(2001)}]{sobol2001global}
\bibinfo{author}{I.~M. Sobol},
\newblock \bibinfo{title}{Global sensitivity indices for nonlinear mathematical models and their monte carlo estimates},
\newblock \bibinfo{journal}{Mathematics and computers in simulation} \bibinfo{volume}{55} (\bibinfo{year}{2001}) \bibinfo{pages}{271--280}.
\bibitem[{Saltelli et~al.(2010)Saltelli, Annoni, Azzini, Campolongo, Ratto, and Tarantola}]{saltelli2010variance}
\bibinfo{author}{A.~Saltelli}, \bibinfo{author}{P.~Annoni}, \bibinfo{author}{I.~Azzini}, \bibinfo{author}{F.~Campolongo}, \bibinfo{author}{M.~Ratto}, \bibinfo{author}{S.~Tarantola},
\newblock \bibinfo{title}{Variance based sensitivity analysis of model output. design and estimator for the total sensitivity index},
\newblock \bibinfo{journal}{Computer physics communications} \bibinfo{volume}{181} (\bibinfo{year}{2010}) \bibinfo{pages}{259--270}.
\bibitem[{MSS(2022)}]{MSS:hisdaily:2022}
\bibinfo{title}{Historical daily records, 2022.}, \bibinfo{howpublished}{\emph{climate-historical-daily} \url{http://www.weather.gov.sg/climate-historical-daily}}, \bibinfo{year}{2022}.
\bibitem[{Meneghetti et~al.(2013)Meneghetti, Ricotta, and Atzori}]{meneghetti2013synthesis}
\bibinfo{author}{G.~Meneghetti}, \bibinfo{author}{M.~Ricotta}, \bibinfo{author}{B.~Atzori},
\newblock \bibinfo{title}{A synthesis of the push-pull fatigue behaviour of plain and notched stainless steel specimens by using the specific heat loss},
\newblock \bibinfo{journal}{Fatigue \& Fracture of Engineering Materials \& Structures} \bibinfo{volume}{36} (\bibinfo{year}{2013}) \bibinfo{pages}{1306--1322}.
\bibitem[{Raman and Thakur(1982)}]{raman1982thermal}
\bibinfo{author}{R.~Raman}, \bibinfo{author}{A.~Thakur},
\newblock \bibinfo{title}{Thermal emissivity of materials},
\newblock \bibinfo{journal}{Applied Energy} \bibinfo{volume}{12} (\bibinfo{year}{1982}) \bibinfo{pages}{205--220}.
\bibitem[{Ritland(1954)}]{ritland1954density}
\bibinfo{author}{H.~N. Ritland},
\newblock \bibinfo{title}{Density phenomena in the transformation range of a borosilicate crown glass},
\newblock \bibinfo{journal}{Journal of the American Ceramic Society} \bibinfo{volume}{37} (\bibinfo{year}{1954}) \bibinfo{pages}{370--377}.
\bibitem[{Sharp and Ginther(1951)}]{sharp1951effect}
\bibinfo{author}{D.~Sharp}, \bibinfo{author}{L.~Ginther},
\newblock \bibinfo{title}{Effect of composition and temperature on the specific heat of glass},
\newblock \bibinfo{journal}{Journal of the American Ceramic Society} \bibinfo{volume}{34} (\bibinfo{year}{1951}) \bibinfo{pages}{260--271}.
\bibitem[{Wang et~al.(2020)Wang, Shan, Shi, Zhang, Cai, and Smith}]{wang2020optically}
\bibinfo{author}{X.~Wang}, \bibinfo{author}{S.~Shan}, \bibinfo{author}{S.~Q. Shi}, \bibinfo{author}{Y.~Zhang}, \bibinfo{author}{L.~Cai}, \bibinfo{author}{L.~M. Smith},
\newblock \bibinfo{title}{Optically transparent bamboo with high strength and low thermal conductivity},
\newblock \bibinfo{journal}{ACS Applied Materials \& Interfaces} \bibinfo{volume}{13} (\bibinfo{year}{2020}) \bibinfo{pages}{1662--1669}.
\bibitem[{Iffat(2015)}]{iffat2015relation}
\bibinfo{author}{S.~Iffat},
\newblock \bibinfo{title}{Relation between density and compressive strength of hardened concrete},
\newblock \bibinfo{journal}{Concrete Research Letters} \bibinfo{volume}{6} (\bibinfo{year}{2015}) \bibinfo{pages}{182--189}.
\bibitem[{De~Schutter and Taerwe(1995)}]{de1995specific}
\bibinfo{author}{G.~De~Schutter}, \bibinfo{author}{L.~Taerwe},
\newblock \bibinfo{title}{Specific heat and thermal diffusivity of hardening concrete},
\newblock \bibinfo{journal}{Magazine of Concrete research} \bibinfo{volume}{47} (\bibinfo{year}{1995}) \bibinfo{pages}{203--208}.
\bibitem[{Kim et~al.(2003)Kim, Jeon, Kim, and Yang}]{kim2003experimental}
\bibinfo{author}{K.-H. Kim}, \bibinfo{author}{S.-E. Jeon}, \bibinfo{author}{J.-K. Kim}, \bibinfo{author}{S.~Yang},
\newblock \bibinfo{title}{An experimental study on thermal conductivity of concrete},
\newblock \bibinfo{journal}{Cement and concrete research} \bibinfo{volume}{33} (\bibinfo{year}{2003}) \bibinfo{pages}{363--371}.
\bibitem[{Olioso(1995)}]{olioso1995simulating}
\bibinfo{author}{A.~Olioso},
\newblock \bibinfo{title}{Simulating the relationship between thermal emissivity and the normalized difference vegetation index},
\newblock \bibinfo{journal}{International Journal of Remote Sensing} \bibinfo{volume}{16} (\bibinfo{year}{1995}) \bibinfo{pages}{3211--3216}.
\bibitem[{Ganguly et~al.(2008)Ganguly, Samanta, Schull, Shabanov, Milesi, Nemani, Knyazikhin, and Myneni}]{ganguly2008generating}
\bibinfo{author}{S.~Ganguly}, \bibinfo{author}{A.~Samanta}, \bibinfo{author}{M.~A. Schull}, \bibinfo{author}{N.~V. Shabanov}, \bibinfo{author}{C.~Milesi}, \bibinfo{author}{R.~R. Nemani}, \bibinfo{author}{Y.~Knyazikhin}, \bibinfo{author}{R.~B. Myneni},
\newblock \bibinfo{title}{Generating vegetation leaf area index earth system data record from multiple sensors. part 2: Implementation, analysis and validation},
\newblock \bibinfo{journal}{Remote Sensing of Environment} \bibinfo{volume}{112} (\bibinfo{year}{2008}) \bibinfo{pages}{4318--4332}.
\bibitem[{Waldemar and Klecha(2015)}]{waldemar2015modeling}
\bibinfo{author}{M.~Waldemar}, \bibinfo{author}{D.~Klecha},
\newblock \bibinfo{title}{Modeling of atmospheric transmission coefficient in infrared for thermovision measurements},
\newblock in: \bibinfo{booktitle}{Proceedings of the Sensor}, pp. \bibinfo{pages}{903--907}.
\bibitem[{MSS(2020)}]{MSS:flirextractor:2020}
\bibinfo{title}{aloisklink/flirextractor v1.0.2: An efficient gplv3-licensed python package for extracting temperature data from flir irt images.}, \bibinfo{howpublished}{\emph{flirextractor} \url{https://github.com/aloisklink/flirextractor}}, \bibinfo{year}{2020}.
\bibitem[{Russell et~al.(2008)Russell, Torralba, Murphy, and Freeman}]{russell2008labelme}
\bibinfo{author}{B.~C. Russell}, \bibinfo{author}{A.~Torralba}, \bibinfo{author}{K.~P. Murphy}, \bibinfo{author}{W.~T. Freeman},
\newblock \bibinfo{title}{Labelme: a database and web-based tool for image annotation},
\newblock \bibinfo{journal}{International journal of computer vision} \bibinfo{volume}{77} (\bibinfo{year}{2008}) \bibinfo{pages}{157--173}.
\bibitem[{Weng and Quattrochi(2018)}]{weng2018urban}
\bibinfo{author}{Q.~Weng}, \bibinfo{author}{D.~A. Quattrochi}, \bibinfo{title}{Urban remote sensing}, \bibinfo{publisher}{CRC press}, \bibinfo{year}{2018}.
\bibitem[{Yang(2021)}]{yang2021urban}
\bibinfo{author}{X.~X. Yang}, \bibinfo{title}{Urban remote sensing: Monitoring, synthesis and modeling in the urban environment}, \bibinfo{publisher}{John Wiley \& Sons}, \bibinfo{year}{2021}.
\bibitem[{Kylili et~al.(2014)Kylili, Fokaides, Christou, and Kalogirou}]{kylili2014infrared}
\bibinfo{author}{A.~Kylili}, \bibinfo{author}{P.~A. Fokaides}, \bibinfo{author}{P.~Christou}, \bibinfo{author}{S.~A. Kalogirou},
\newblock \bibinfo{title}{Infrared thermography (irt) applications for building diagnostics: A review},
\newblock \bibinfo{journal}{Applied Energy} \bibinfo{volume}{134} (\bibinfo{year}{2014}) \bibinfo{pages}{531--549}.
\bibitem[{Floridi et~al.(2018)Floridi, Cowls, Beltrametti, Chatila, Chazerand, Dignum, Luetge, Madelin, Pagallo, Rossi et~al.}]{floridi2018ai4people}
\bibinfo{author}{L.~Floridi}, \bibinfo{author}{J.~Cowls}, \bibinfo{author}{M.~Beltrametti}, \bibinfo{author}{R.~Chatila}, \bibinfo{author}{P.~Chazerand}, \bibinfo{author}{V.~Dignum}, \bibinfo{author}{C.~Luetge}, \bibinfo{author}{R.~Madelin}, \bibinfo{author}{U.~Pagallo}, \bibinfo{author}{F.~Rossi}, et~al.,
\newblock \bibinfo{title}{Ai4people—an ethical framework for a good ai society: Opportunities, risks, principles, and recommendations},
\newblock \bibinfo{journal}{Minds and machines} \bibinfo{volume}{28} (\bibinfo{year}{2018}) \bibinfo{pages}{689--707}.
\bibitem[{Tom~Yeh et~al.(2017)}]{tom2017designing}
\bibinfo{author}{M.~Tom~Yeh}, et~al.,
\newblock \bibinfo{title}{Designing a moral compass for the future of computer vision using speculative analysis},
\newblock in: \bibinfo{booktitle}{Proceedings of the IEEE Conference on Computer Vision and Pattern Recognition Workshops}, pp. \bibinfo{pages}{64--73}.
\bibitem[{lan(2014)}]{lane_stodden_bender_nissenbaum_2014}
\bibinfo{title}{Privacy, Big Data, and the Public Good: Frameworks for Engagement}, \bibinfo{publisher}{Cambridge University Press}, \bibinfo{year}{2014}.
\bibitem[{McCallister(2010)}]{mccallister2010guide}
\bibinfo{author}{E.~McCallister}, \bibinfo{title}{Guide to protecting the confidentiality of personally identifiable information}, volume \bibinfo{volume}{800}, \bibinfo{publisher}{Diane Publishing}, \bibinfo{year}{2010}.

\end{thebibliography}

\end{document}